\documentclass{article}
\usepackage{BGNet-arxiv}

\usepackage{times}
\usepackage{soul}
\usepackage{url}
\usepackage[utf8]{inputenc}
\usepackage[small]{caption}
\usepackage{graphicx}
\usepackage{amsmath}
\usepackage{amsthm}
\usepackage{amssymb}
\usepackage{booktabs}
\usepackage{algorithm}
\usepackage{algorithmic}
\usepackage{multirow}
\usepackage{color}
\usepackage{colortbl}
\usepackage[breaklinks=true,colorlinks,bookmarks=false]{hyperref}

\usepackage[capitalize]{cleveref}
\crefname{section}{Sec.}{Secs.}
\Crefname{section}{Section}{Sections}
\Crefname{table}{Table}{Tables}
\crefname{table}{Tab.}{Tabs.}
\graphicspath{{./img/}}
\DeclareGraphicsExtensions{.jpg,.pdf,.png}

\def\ie{\emph{i.e.}}
\def\eg{\emph{e.g.}}

\definecolor{ours}{gray}{.95}

\title{Boundary-Guided Camouflaged Object Detection}

\author{
Yujia Sun$^1$\and
Shuo Wang$^2$\and
Chenglizhao Chen$^{3}$\and
Tian-Zhu Xiang$^4$\footnote{Corresponding Author (tianzhu.xiang19@gmail.com).}
\affiliations
$^1$School of Computer Science, Inner Mongolia University, China \\
$^2$ETH Zurich, Switzerland  \\
$^3$College of Computer Science and Technology, China University of Petroleum, China\\
$^4$Inception Institute of Artificial Intelligence, UAE
}

\begin{document}

\maketitle

\begin{abstract}
  Camouflaged object detection (COD), segmenting objects that are elegantly blended into their surroundings, is a valuable yet challenging task. Existing deep-learning methods often fall into the difficulty of accurately identifying the camouflaged object with complete and fine object structure. To this end, in this paper, we propose a novel boundary-guided network (\textit{BGNet}) for camouflaged object detection. Our method explores valuable and extra object-related edge semantics to guide representation learning of COD, which forces the model to generate features that highlight object structure, thereby promoting camouflaged object detection of accurate boundary localization. Extensive experiments on three challenging benchmark datasets demonstrate that our \textit{BGNet} significantly outperforms the existing 18 state-of-the-art methods under four widely-used evaluation metrics. Our code is publicly available at:~\href{https://github.com/thograce/BGNet}{https://github.com/thograce/BGNet}.
\end{abstract}

\section{Introduction}

Camouflage is an important defense mechanism in nature that helps certain species hide in the surroundings to protect themselves from their predators, through concealment by means of materials, coloration or illumination, or self-disguise as something else, such as imitating the appearance, colors, or patterns of the environment and disruptive coloration~\cite{price2019background}. This mechanism also affects human life, such as art, culture and design (\textit{e.g.}, camouflaged uniforms)~\cite{stevens2009animal}. Recently, identifying a camouflaged object from its background, namely camouflaged object detection (COD), has attracted increasing research interest from computer vision community. It has promising prospects for facilitating various valuable applications in different fields, ranging from animal conservation, \textit{e.g.}, species discovery~\cite{perez2012early} and animal monitoring, to vision-related areas, including image synthesis~\cite{sinet}, medical image analysis~\cite{pranet}, and search-and-rescue. However, COD is a very challenging task due to the nature of camouflage, that is, the high intrinsic similarities between candidate objects and chaotic background, which make it difficult to spot camouflaged objects for humans and machines.

\begin{figure}[bt]
    \centering
	\includegraphics[width=0.95\linewidth]{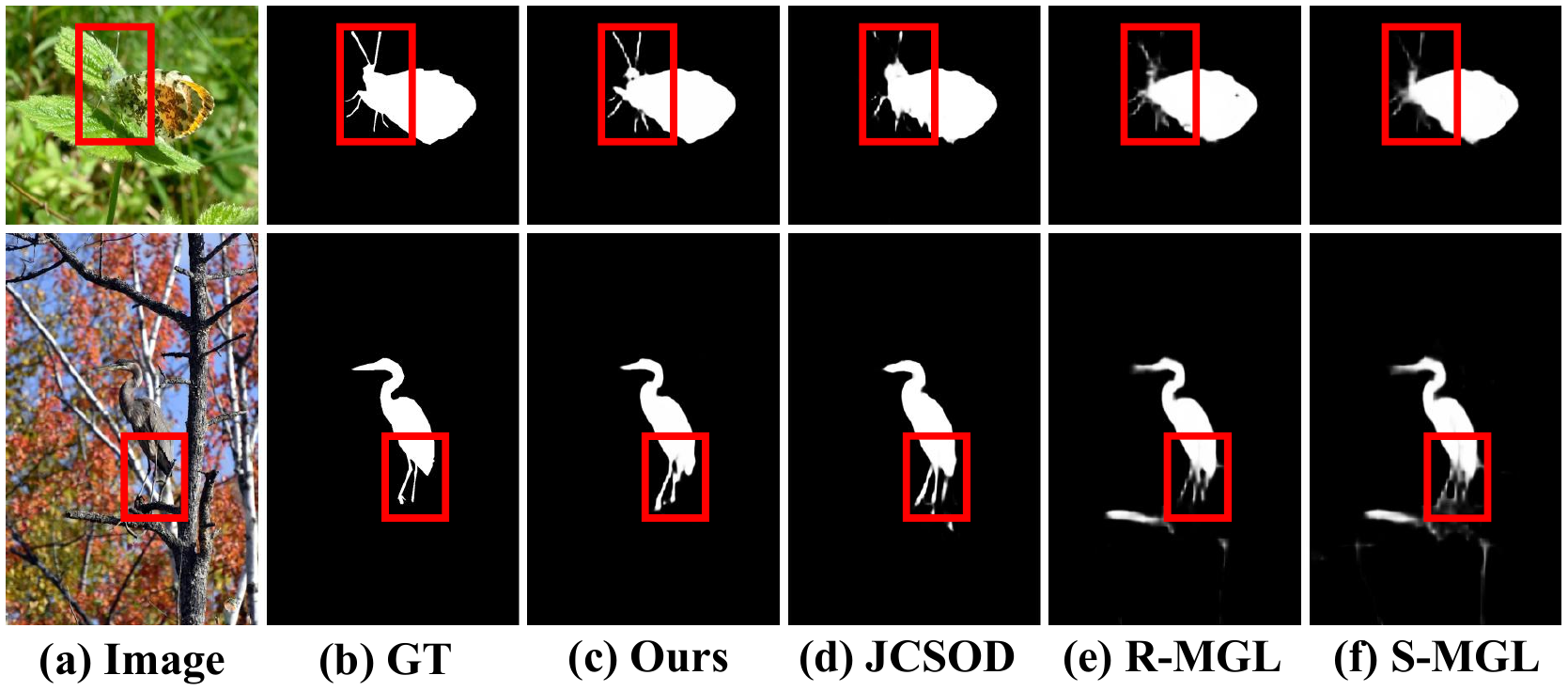}
	\caption{Visual examples of camouflaged object detection in some challenging scenarios. Compared with the recently proposed deep learning-based methods (\eg, JCSOD, S-MGL, and R-MGL), our method, incorporating \textit{\textbf{object-related edge semantics}}, can effectively distinguish the object structure and details, and produce accurate predictions with intact boundaries, as shown in red boxes. 
	}\label{fig:firstpage}%\vspace{-3mm}
\end{figure}

To tackle this issue, numerous deep learning-based methods have been proposed for camouflaged object detection and have shown great potentials. These methods can be broadly divided into three types. One is to design targeted network modules/architectures to effectively explore discriminative camouflaged-object features for COD, such as C$^2$FNet~\cite{c2fnet}, UGTR~\cite{ugtr}. One is to incorporate some auxiliary tasks into the joint learning/multi-task learning frameworks, such as classification task~\cite{camo}, edge extraction~\cite{mgl}, salient object detection~\cite{jcsod}, and camouflaged object ranking~\cite{lsr}. This kind of method can excavate valuable and extra cues from the shared features to significantly enhance the feature representation for COD. Another is the bio-inspired method, which mimics the behavior process of predators in nature or human visual psychological patterns to design the networks, such as SINet~\cite{sinet,sinet2}, MirrorNet~\cite{mirrornet}, and PFNet~\cite{pfnet}.

Although these recently proposed methods have made significant progress, there are still some major issues. Existing methods are often difficult to effectively and completely identify the structure and details of objects due to the edge disruption or body outline disguising, thus providing unsatisfactory predictions with coarse/incomplete object boundaries. As shown in Fig.~\ref{fig:firstpage}, the ambiguities, caused by the high similarities between the butterfly head/heron legs and its background surroundings, make the features extracted by the recently state-of-the-art JCSOD~\cite{jcsod} and MGL~\cite{mgl} (including single-stage version and recurrent version) models indistinguishable. Thus these models are incapable of recovering the boundary details of butterfly head and heron legs.
As mentioned in~\cite{zhao2019egnet}, the edge prior is widely used as an effective auxiliary cue, which benefits the preservation of object structure, yet has been barely studied for COD. Empirically, it is necessary to study how to enhance object-related edge visibility for facilitating the feature learning of COD. To our best knowledge, the mutual graph learning model (MGL)~\cite{mgl} is the first to explicitly exploit the edge information to improve the performance of COD. However, MGL encodes edge features into graph convolutional networks along with object features and boosts feature representation through graph interaction modules. It can be seen that MGL is a sophisticated model, which inevitably increases model complexity and suffers from the heavy computational burden. In addition, it is noted that, despite the introduction of edge cues, MGL still loses some fine boundary-related details and thus weakens the performance of COD. As shown in the 2$^{nd}$ row of Fig.~\ref{fig:firstpage}, the MGL models lose many details about heron legs and introduces some obvious background noises in prediction results.

To this end, in this paper, we propose a novel boundary-guided network (\textit{BGNet}), which explicitly employs edge semantic to enhance the performance of camouflaged object detection.
Firstly, we design a simple yet effective edge-aware module (EAM), which integrates the low-level local edge information and high-level global location information to explore edge semantics related to object boundaries under explicitly boundary supervision.
Then, the edge-guidance feature module (EFM) is introduced to incorporate the edge features with the camouflaged object features at various layers to guide the representation learning of COD. The EFM module can enforce the network to pay more attention to the object structure and details. 
After that, the multi-level fused features are aggregated gradually from top to bottom to predict camouflaged objects. To enhance the feature representation, we build a context aggregation module (CAM), which mines and aggregates multi-scale contextual semantics by a series of atrous convolutions to produce features with stronger and more effective representations.
Benefiting from the well-designed modules, the proposed \textit{BGNet} predicts camouflaged objects with fine object structure and boundaries. Note that, compared with MGL, we design a simpler but more effective edge extraction module to excavate accurate object boundary semantics, and then guide the feature representation learning of camouflaged objects by means of the proposed EFM and CAM. Besides, our method achieves more accurate object localization and stronger preservation of object structure.
To sum up, our main contributions are as follows:
\begin{itemize}
    \item For the COD task, we propose a novel boundary-guided network, \ie, \textit{BGNet}, which excavates and integrates boundary-related edge semantics to boost the performance of camouflaged object detection.
    \item We carefully design the edge-guidance feature module (EFM) and context aggregation module (CAM) to enhance boundary semantics and explore valuable and powerful feature representation for COD.
    \vspace{-0.5mm}
\end{itemize}

\section{Related Work}
\label{sec:related}

\paragraph{Datasets.} To boost deep learning-based camouflaged object detection, some annotation datasets are proposed. \cite{camo} constructed the first camouflaged object dataset, namely CAMO, including 1,250 camouflaged images covering eight categories. \cite{sinet} collected a large-scale challenging dataset, called COD10K, which contains 10,000 images covering 78 camouflaged object categories, with high-quality and hierarchical annotations. Recently, to support localization and ranking of camouflaged objects, \cite{lsr} proposed a ranking-based testing dataset, named NC4K, which contains 4,121 images with extra localization annotation and ranking annotation.

\paragraph{Camouflaged object detection.} In recent years, camouflaged object detection has attracted increasing attention in the computer vision community~\cite{ZoomNet,videoCod}. Since the release of large-scale data sets (\textit{e.g.}, CAMO and COD10K), numerous deep learning-based camouflaged object detection models have been proposed and achieved great progress. These methods can be roughly divided into three types. 
The first type of method is to design advanced network modules/architectures to explore discriminative camouflaged features for COD. \cite{c2fnet} designed an attention-induced cross-level fusion module and a dual-branch global context module to enhance feature representation. \cite{ugtr} incorporated the Bayesian learning into transformer-based reasoning, which can leverage both deterministic and probabilistic information for COD.
The second type of method is to combine some auxiliary tasks into the joint learning frameworks to boost the performance of COD. \cite{camo} proposed an anabranch network that performs an auxiliary classification network to help camouflaged object segmentation. \cite{mgl} exploited edge extraction as an auxiliary task and incorporated it into mutual graph learning for COD. \cite{jcsod} presented a joint salient object detection and camouflaged object detection network to enhance the detection ability of both tasks. To segment and rank camouflaged objects, \cite{lsr} designed a ranking-based COD model in a joint learning framework that can mutually boost the performance of each another.
The last one is the bio-inspired method, which is inspired by the behavior process of predators in nature or human visual psychological patterns. \cite{sinet} presented a search identification network to gradually locate and search for the camouflaged object, inspired by the process of wild predators discovering prey. \cite{pfnet} proposed a positioning and focus network by mimicking the detection and identification stages of predation.

\section{Proposed Method}
\label{sec:method}

\subsection{Overall Architecture}
\begin{figure}[t]
    \centering
    \includegraphics[width=1\linewidth]{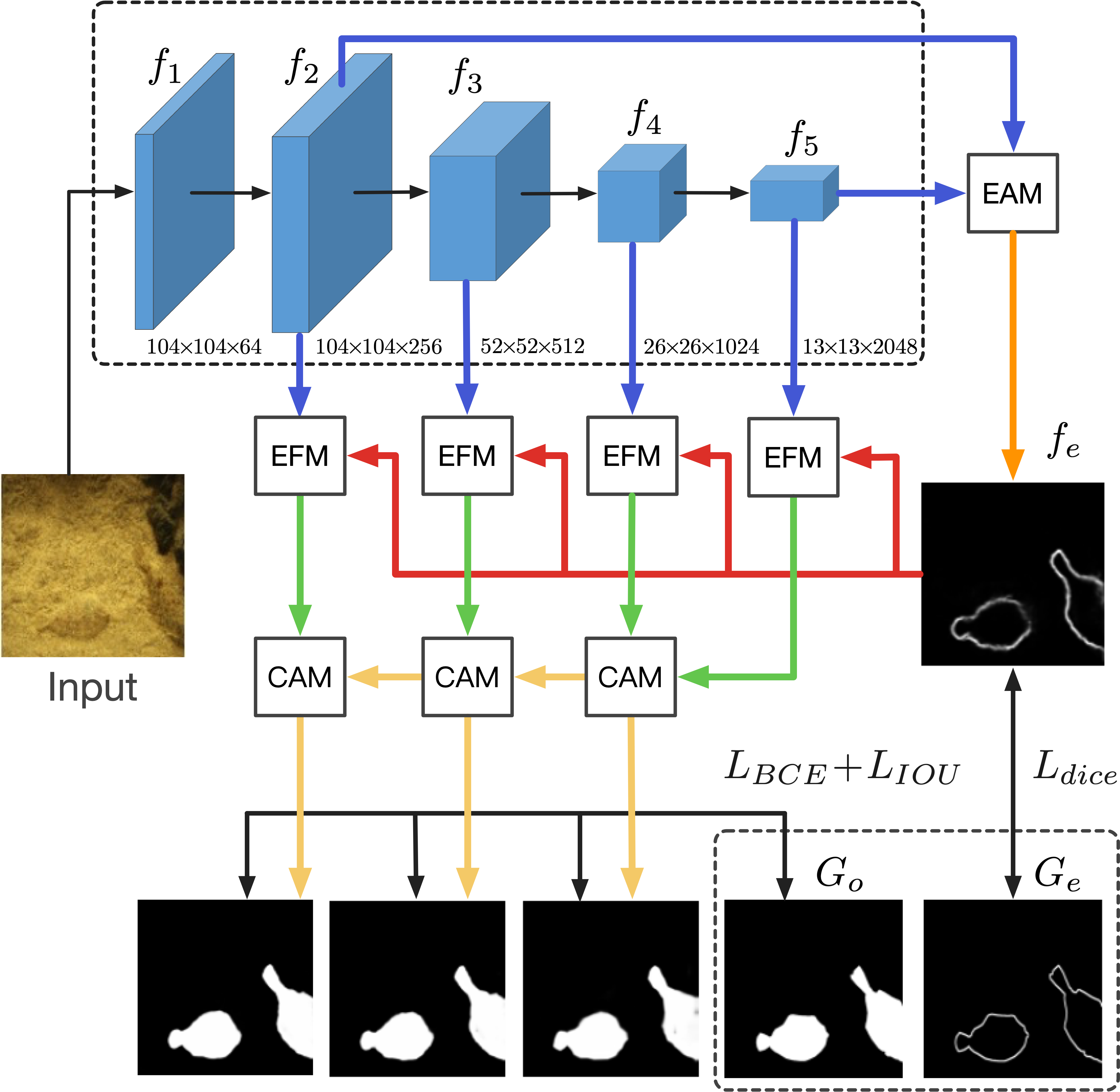} 
	\caption{The overall architecture of the proposed \textit{BGNet}, which consists of three key components, \ie, edge-aware module (EAM), edge-guidance feature module (EFM) and context aggregation module (CAM). See \cref{sec:method} for details. 
	}\vspace{-6pt}
    \label{fig:net_overall}
\end{figure}

The overall architecture of the proposed \textit{BGNet} is illustrated in Fig.~\ref{fig:net_overall}. Specifically, we adopt Res2Net-50~\cite{res2net} as our backbone network to extract multi-level features from the input image, \ie,  $f_i~(i=1,2,\dots,5)$. 
Then, an edge-aware module (EAM) is applied to excavate object-related edge semantics from the low-level features, which contain local edge details ($f_2$), and the high-level features, which contain global location information ($f_5$) under object boundary supervision. 
Following multiple edge-guidance feature modules (EFM) are leveraged to integrate the edge cues from EAM with multi-level backbone features ($f_2$-$f_5$) at each level to guide feature learning, which enhances boundary representation. 
Finally, multiple context aggregation modules (CAM) are employed to progressively aggregate multi-level fused features in a top-down manner and discover camouflaged objects. In testing, we select the prediction of the last CAM as the final result. Noted that, we do not adopt the $f_1$ backbone feature because it is too close to the input with much redundant information and a small receptive field.

\subsection{Edge-aware Module}

A good edge prior can benefit object detection in both segmentation and localization~\cite{zhang2017amulet,zhao2019egnet}. Although low-level features contain rich edge details, they also introduce many non-object edges. Thus, high-level semantic or location information is needed to facilitate the exploration of camouflaged object-related edge features. 
In this module, we incorporate the low-level feature ($f_2$) and the high-level feature ($f_5$) to model the object-related edge information, as shown in Fig.~\ref{fig:net_eam}.
Specifically, two 1$\times$1 convolution layers are first used to change the channels of $f_2$ and $f_5$ to 64 ($f_2^{'}$) and 256 ($f_5^{'}$), respectively. Then we integrate the feature $f_2^{'}$ and the up-sampled $f_5^{'}$ by concatenation operation. Finally, we obtain the edge feature $f_e$ through two 3$\times$3 convolution layers and one 1$\times$1 convolution layer followed by the Sigmoid function. EAM is a simple yet effective module to extract specific edge features. As shown in Fig.~\ref{fig:edge}, EAM perfectly learns the object boundary-related edge semantics.

\begin{figure}[bt]
    \centering
	\includegraphics[width=1\linewidth]{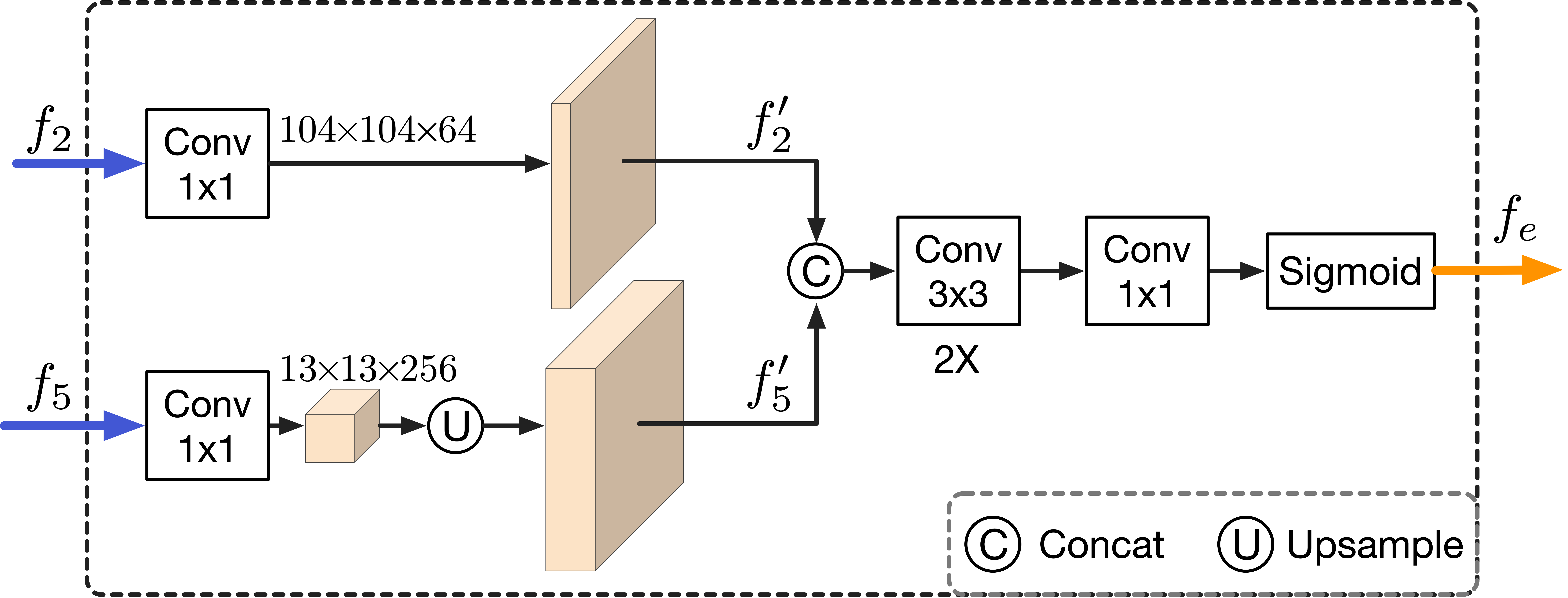} 
	\caption{Illustration of the edge-aware module (EAM), which is designed to excavate camouflaged object-related edge features.}
    \label{fig:net_eam}
\end{figure}

\subsection{Edge-guidance Feature Module}

The edge-guidance feature module (EFM) is designed to inject boundary-related edge cues into the representation learning to enhance the feature representation with object structure semantics. As is known to all, different feature channels often contain differentiated semantics. Thus, to achieve good integration and obtain a powerful representation, we introduce a local channel attention mechanism to explore cross-channel interaction and mine the critical cues between channels.

\begin{figure}
    \centering{
	\includegraphics[width=1\linewidth]{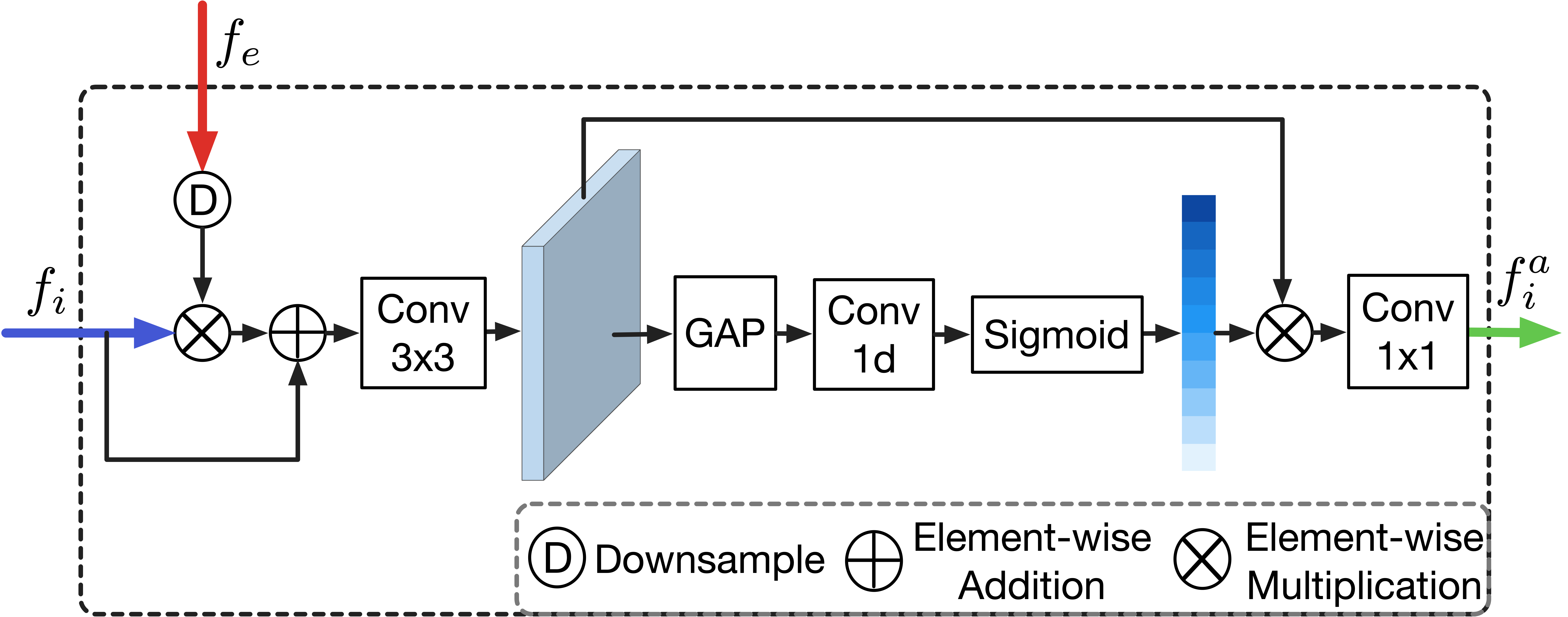}
	\caption{Illustration of the edge-guidance feature module (EFM) adopted to integrate edge cues to guide the representation learning.}
    \label{fig:efm}
    }\vspace{-5pt}
\end{figure}

As shown in Fig.~\ref{fig:efm}, given the input feature $f_i~(i\in\{2,3,...,5\})$ and the edge feature $f_e$, we first perform the element-wise multiplication between them with an additional skip-connection and a 3$\times$3 convolution to obtain the initial fused features $f_i^e$, which can be denoted as: 
\begin{equation}
f_{i}^{e} = \mathbf{F}_{conv}( (f_{i}\otimes D(f_e))\oplus f_{i} ),
\end{equation}
where $D$ denotes down-sampling and $\mathbf{F}_{conv}$ is 3$\times$3 convolution. $\otimes$ is element-wise multiplication and $\oplus$ is element-wise addition.
To enhance feature representation, inspired by~\cite{eca}, we introduce local attention to explore critical feature channels. Specifically, we aggregate the convolution features ($f_i^e$) using a channel-wise global average pooling (GAP). 
Then we obtain the corresponding channel attention (weight) by the 1D convolutions followed by a Sigmoid function. Unlike fully-connected operations, which capture dependencies across all channels but show high complexity, we explore local cross-channel interaction and learn each attention in a local manner, \eg, only consider its $k$ neighbors of every channel. 
After that, we multiply the channel attention with the input feature $f_i^e$ and reduce the channels by 1$\times$1 convolution layer to obtain the final output $f_i^a$, \ie, 
\begin{equation}
f_{i}^{a} = \mathbf{F}_{conv1}(\sigma(\mathbf{F}_{1D}^k(GAP(f_{i}^{e}))) \otimes f_{i}^{e}),
\end{equation}
where $\mathbf{F}_{conv1}$ is 1$\times$1 convolution, $\mathbf{F}_{1D}^k$ is 1D convolution with kernel size $k$ and $\sigma$ denotes Sigmoid function. The kernel size $k$ can be set adaptively as $k = \left| (1+log_{2}(C))/2 \right|_{odd}$, where $|*|_{odd}$ denotes the nearest odd number and $C$ is the channels of $f_{i}^{e}$. The kernel size is proportional to channel dimension. 
Obviously, the proposed attention strategy can highlight critical channels and suppress redundant channels or noises, thereby enhancing semantic representation.

\begin{figure}[bt]
    \centering{
	\includegraphics[width=1\linewidth]{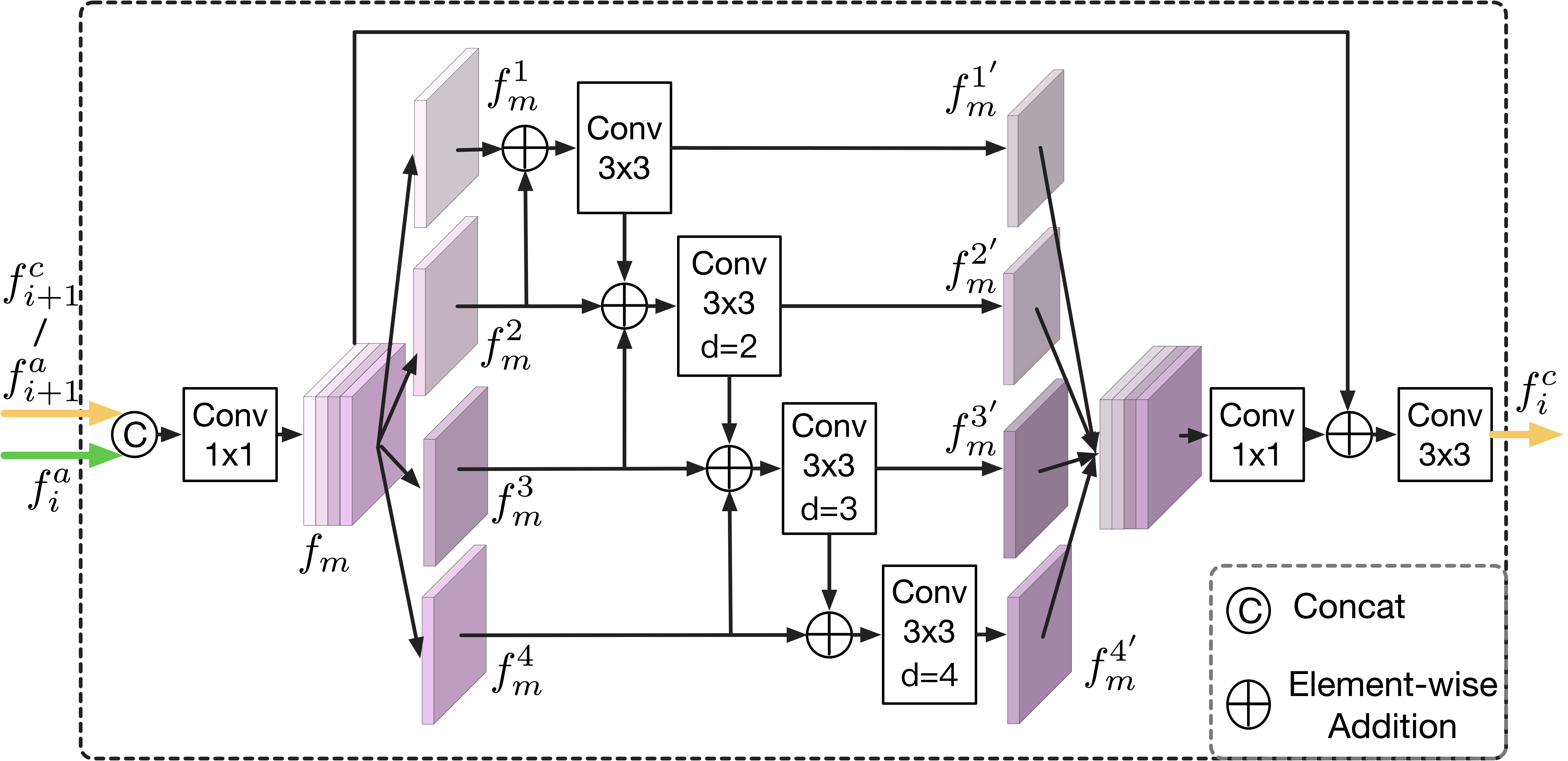}
	\caption{Illustration of the context aggregation module (CAM) used to mine contextual semantics for enhancing feature representation.}
    \label{fig:cam}\vspace{-4pt}
    }
\end{figure}

\subsection{Context Aggregation Module}

To integrate multi-level fused features for camouflaged object prediction, we design a context aggregation module (CAM) to mine contextual semantics for boosting object detection, as shown in Fig.~\ref{fig:cam}. Unlike global context modules in BBS-Net~\cite{fan2020bbs}, which only exploits several independent parallel branches to extract features of different scales separately without considering the semantic correlation among various branches~\cite{edn}, CAM takes cross-scale interaction into account to boost feature representation. 
Taking $f_4^a$ and $f_{5}^a$ as an example, we first upsample $f_{5}^a$ and concatenate them followed by a 1$\times$1 convolution layer to obtain the initial aggregated feature $f_m$. Next, we evenly divide $f_m$ into four feature maps ($f_m^1, f_m^2, f_m^3, f_m^4$) along channel dimension, and then perform cross-scale interaction learning, that is, to integrate the features of adjacent branches to extract multi-scale contextual features by a series of atrous convolutions. It can be formulated as:
\begin{equation}
f_m^{j{'}} = \mathbf{F}_{conv}^{n_j} ( f_m^{{j-1}{'}} \oplus f_m^j \oplus f_m^{j+1} ), ~j\in \{1,2,3,4\}, 
\end{equation}
where $\mathbf{F}_{conv}^{n_j}$ indicates a $3\times3$ atrous convolution with a dilation rate of $n_j$. In our experiments, we set $n_j = \{1, 2, 3, 4\}$. Besides, for $i=1$, there is only $f_m^1$ and $f_m^2$; for $i=4$, there is only $f_m^4$ and $f_m^{3'}$. Then, we concatenate these four multi-scale features $f_m^{j{'}}$ followed by a 1$\times$1 convolution, a residual connection and a 3$\times$3 convolution, which can be denoted as:
\begin{equation}
f_i^c = \mathbf{F}_{conv} ( \mathbf{F}_{conv1} ( [f_m^{j{'}}] ) \oplus f_m ),
\end{equation}
where $[*]$ is concatenation operation, and $f_i^c$ is output of CAM. Noted that, for $i=\{2,3\}$, the output of the previous CAM ($f_{i+1}^c$) will be used together with $f_{i}^a$ as the input of the next CAM to obtain the $f_i^c$. With another 1$\times$1 convolution to change the channel number of feature $f_i^c$, we can obtain the prediction $P_i$~($i\in\{2,3,4\}$ of camouflaged objects.

\subsection{Loss Function}
Our model has two kinds of supervisions: camouflaged object mask ($G_o$) and camouflaged object edge ($G_e$). For mask supervision, we employ the weighted binary cross-entropy loss (${L}_{BCE}^{w}$) and weighted IOU loss (${L}_{IOU}^{w}$)~\cite{f3net}, which pay more attention to hard pixels instead of assigning all pixels equal weights. 
For edge supervision, we adopt the dice loss (${L}_{dice}$)~\cite{transparent} to deal with the strong imbalance between positive and negative samples. 
Noted the mask supervision is conducted on three camouflaged object predictions ($P_i, i\in\{2,3,4\}$) from CAM.
Thus, the total loss is defined as: ${L}_{total}=\sum_{i=2}^{4}({L}_{BCE}^{w}(P_i, G_o) + {L}_{IOU}^{w}(P_i, G_o)) + \lambda {L}_{dice}(P_e, G_e)$, where $\lambda$ is a trade-off parameter and set $\lambda=3$ in our experiments, $P_e$ is the prediction of camouflaged object edges.

\begin{table*}[th]
%\vspace{-3pt}
\resizebox{\textwidth}{!}{
\begin{tabular}{l|r|cccc|cccc|cccc}
\toprule
\multirow{2}{*}{Method} & \multirow{2}{*}{Pub./Year} & \multicolumn{4}{c|}{CAMO-Test} & \multicolumn{4}{c|}{COD10K-Test} & \multicolumn{4}{c}{NC4K} \\ \cline{3-14} 
 &  & $S_\alpha\uparrow$ & $E_\phi\uparrow$ & $F_\beta^w\uparrow$ & $\mathcal{M}\downarrow$ & $S_\alpha\uparrow$ & $E_\phi\uparrow$ & $F_\beta^w\uparrow$ & $\mathcal{M}\downarrow$ & $S_\alpha\uparrow$ & $E_\phi\uparrow$ & $F_\beta^w\uparrow$ & $\mathcal{M}\downarrow$ \\ \midrule
PoolNet & CVPR'19 & 0.730 & 0.746 & 0.575 & 0.105 & 0.740 & 0.776 & 0.506 & 0.056 & 0.785 & 0.814 & 0.635 & 0.073 \\
EGNet & ICCV'19 & 0.732 & 0.800 & 0.604 & 0.109 & 0.736 & 0.810 & 0.517 & 0.061 & 0.777 & 0.841 & 0.639 & 0.075 \\
SRCN & ICCV'19 & 0.779 & 0.797 & 0.643 & 0.090 & 0.789 & 0.817 & 0.575 & 0.047 & 0.830 & 0.854 & 0.698 & 0.059 \\
F$^3$Net & AAAI'20 & 0.711 & 0.741 & 0.564 & 0.109 & 0.739 & 0.795 & 0.544 & 0.051 & 0.780 & 0.824 & 0.656 & 0.070 \\
ITSD & CVPR'20 & 0.750 & 0.779 & 0.610 & 0.102 & 0.767 & 0.808 & 0.557 & 0.051 & 0.811 & 0.844 & 0.680 & 0.064 \\
CSNet & ECCV'20 & 0.771 & 0.795 & 0.642 & 0.092 & 0.778 & 0.809 & 0.569 & 0.047 & 0.750 & 0.773 & 0.603 & 0.088 \\
MINet & CVPR'20 & 0.748 & 0.791 & 0.637 & 0.090 & 0.770 & 0.832 & 0.608 & 0.042 & 0.812 & 0.862 & 0.720 & 0.056 \\
UCNet & CVPR'20 & 0.739 & 0.787 & 0.640 & 0.094 & 0.776 & 0.857 & 0.633 & 0.042 & 0.811 & 0.871 & 0.729 & 0.055 \\
PraNet & MICCAI'20 & 0.769 & 0.825 & 0.663 & 0.094 & 0.789 & 0.861 & 0.629 & 0.045 & 0.822 & 0.876 & 0.724 & 0.059 \\
BASNet & arxiv'21 & 0.749 & 0.796 & 0.646 & 0.096 & 0.802 & 0.855 & 0.677 & 0.038 & 0.817 & 0.859 & 0.732 & 0.058 \\
\midrule
SINet & CVPR'20 & 0.745 & 0.804 & 0.644 & 0.092 & 0.776 & 0.864 & 0.631 & 0.043 & 0.808 & 0.871 & 0.723 & 0.058 \\
PFNet & CVPR'21 & 0.782 & 0.841 & 0.695 & \textcolor{blue}{0.085} & 0.800 & 0.877 & 0.660 & 0.040 & 0.829 & 0.887 & 0.745 & 0.053 \\ 
S-MGL & CVPR'21 & 0.772 & 0.806 & 0.664 & 0.089 & 0.811 & 0.844 & 0.654 & 0.037 & 0.829 & 0.862 & 0.731 & 0.055 \\ 
R-MGL & CVPR'21 & 0.775 & 0.812 & 0.673 & 0.088 & \textcolor{blue}{0.814} & 0.851 & 0.666 & \textcolor{green}{0.035} & 0.833 & 0.867 & 0.739 & 0.053 \\
UGTR & ICCV'21 & 0.784 & 0.821 & 0.683 & 0.086 & \textcolor{green}{0.817} & 0.852 & 0.665 & \textcolor{blue}{0.036} & 0.839 & 0.874 & 0.746 & 0.052 \\ 
LSR & CVPR'21 & 0.787 & 0.838 & 0.696 & \textcolor{green}{0.080} & 0.804 & 0.880 & 0.673 & 0.037 & \textcolor{blue}{0.840} & 0.895 & \textcolor{blue}{0.766} & \textcolor{blue}{0.048} \\ 
C$^2$FNet & IJCAI'21 & \textcolor{blue}{0.796} & \textcolor{blue}{0.854} & \textcolor{blue}{0.719} & \textcolor{green}{0.080} & 0.813 & \textcolor{green}{0.890} & \textcolor{green}{0.686} & \textcolor{blue}{0.036} & 0.838 & \textcolor{blue}{0.897} & 0.762 & 0.049 \\
JCSOD & CVPR'21 & \textcolor{green}{0.800} & \textcolor{green}{0.859} & \textcolor{green}{0.728} & \textcolor{red}{0.073} & 0.809 & \textcolor{blue}{0.884} & \textcolor{blue}{0.684} & \textcolor{green}{0.035} & \textcolor{green}{0.841} & \textcolor{green}{0.898} & \textcolor{green}{0.771} & \textcolor{green}{0.047} \\ 
\rowcolor{ours}
BGNet~(Ours) & {IJCAI'22} & \textcolor{red}{0.812} & \textcolor{red}{0.870} & \textcolor{red}{0.749} & \textcolor{red}{0.073} & \textcolor{red}{0.831} & \textcolor{red}{0.901} & \textcolor{red}{0.722} & \textcolor{red}{0.033} & \textcolor{red}{0.851} & \textcolor{red}{0.907} & \textcolor{red}{0.788} & \textcolor{red}{0.044} \\ \bottomrule
\end{tabular}}
\caption{Quantitative comparison with state-of-the-art methods for COD on three benchmarks using four widely used evaluation metrics (\ie, $S_{\alpha}$, $E_{\phi}$, $F_{\beta}^w$, and $\mathcal{M}$). ``$\uparrow$" / ``$\downarrow$" indicates that larger/smaller is better. Top three results are highlighted in \textcolor{red}{red}, \textcolor{green}{green} and \textcolor{blue}{blue}.}
\label{tab1}
\end{table*}

\begin{table*}[ht]
\resizebox{\textwidth}{!}{
\begin{tabular}{c|l|cccc|cccc|cccc}
\hline
\multicolumn{1}{c|}{\multirow{2}{*}{M}} &
  \multicolumn{1}{c|}{\multirow{2}{*}{Method}} &
  \multicolumn{4}{c|}{CAMO-Test} &
  \multicolumn{4}{c|}{COD10K-Test} &
  \multicolumn{4}{c}{NC4K} \\ \cline{3-14} 
\multicolumn{1}{c|}{} &
  \multicolumn{1}{c|}{} &
  $S_\alpha\uparrow$ & $E_\phi\uparrow$ & $F_\beta^w\uparrow$ & $\mathcal{M}\downarrow$ & $S_\alpha\uparrow$ &
  $E_\phi\uparrow$ & $F_\beta^w\uparrow$ & $\mathcal{M}\downarrow$ &
  $S_\alpha\uparrow$ & $E_\phi\uparrow$ & $F_\beta^w\uparrow$ & $\mathcal{M}\downarrow$ \\ \hline
a & B     & 0.799 & 0.858 & 0.726 & 0.080 & 0.823 & 0.893 & 0.702 & 0.035 & 0.845 & 0.897 & 0.773 & 0.048 \\ %\cline{1-2}
b & B+CAM   & 0.809 & 0.864 & 0.738 & \textbf{0.073} & 0.828 & 0.897 & 0.713 & 0.034 & \textbf{0.851} & 0.901 & 0.783 & 0.045 \\ %\cline{1-2}
c & B+EAM+EFM w/o LCA   & 0.805 & 0.866 & 0.733 & 0.075 & 0.828 & 0.898 & 0.713 & 0.034 & 0.848 & 0.902 & 0.780 & 0.046 \\ %\cline{1-2}
d & B+EAM+EFM   & 0.808 & 0.867 & 0.740 & 0.075 & \textbf{0.831} & 0.898 & 0.717 & \textbf{0.033} & 0.849 & 0.903 & 0.781 & 0.046 \\ %\cline{1-2}
e & Ours & \textbf{0.812} & \textbf{0.870} & \textbf{0.749} & \textbf{0.073} & \textbf{0.831} & \textbf{0.901} & \textbf{0.722} & \textbf{0.033} & \textbf{0.851} & \textbf{0.907} & \textbf{0.788} & \textbf{0.044} \\ \hline
\end{tabular}}
\caption{Quantitative evaluation for ablation studies on three datasets. The best results are highlighted in \textbf{Bold}. B: baseline. M: model.}
\label{tab:ablation}
% \vspace{-5pt}
\end{table*}

\begin{figure*}[ht]
    \centering{
    }
	\includegraphics[width=0.95\textwidth]{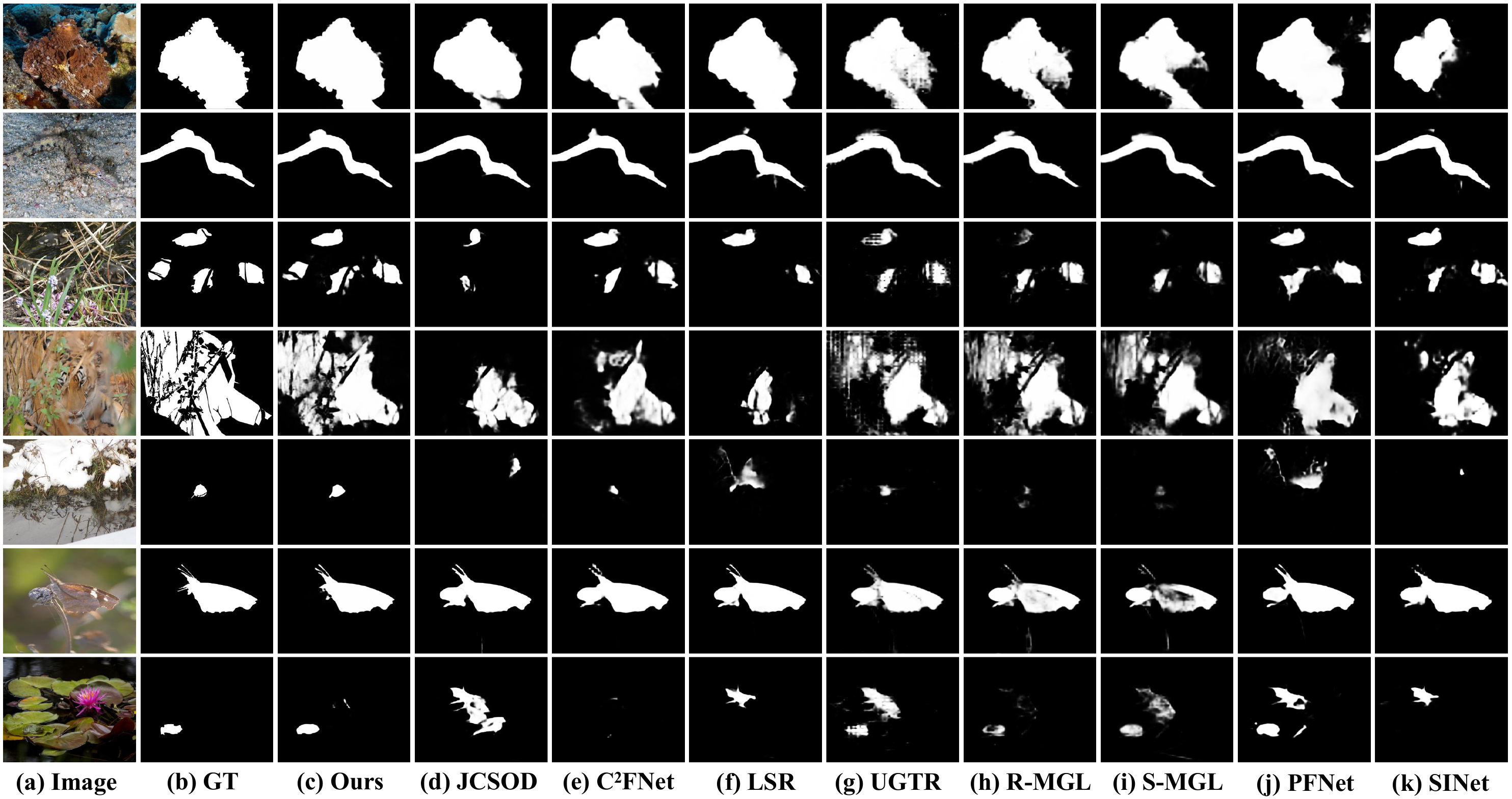}
	\caption{Visual comparison of the proposed model with eight state-of-the-art COD methods. Obviously, our method is capable of accurately segmenting various camouflaged objects with more clear boundaries.}
    \label{fig:results}
    % \vspace{-1.5mm}
\end{figure*}

\begin{table*}[ht]
\centering
% \vspace{-3pt}
\resizebox{\textwidth}{!}{
\begin{tabular}{l|cccc|cccc|cccc}
\hline
% \multicolumn{1}{c|}{\multirow{2}{*}{M}} &
  \multicolumn{1}{c|}{\multirow{2}{*}{Method}} &
  \multicolumn{4}{c|}{CAMO-Test} &
  \multicolumn{4}{c|}{COD10K-Test} &
  \multicolumn{4}{c}{NC4K} \\ \cline{2-13} 
% \multicolumn{1}{c|}{} &
  \multicolumn{1}{c|}{} &
  $S_\alpha\uparrow$ & $E_\phi\uparrow$ & $F_\beta^w\uparrow$ & $\mathcal{M}\downarrow$ & $S_\alpha\uparrow$ &
  $E_\phi\uparrow$ & $F_\beta^w\uparrow$ & $\mathcal{M}\downarrow$ &
  $S_\alpha\uparrow$ & $E_\phi\uparrow$ & $F_\beta^w\uparrow$ & $\mathcal{M}\downarrow$ \\ \hline
 $f_1$+$f_5$     & 0.807 & 0.865 & 0.737 & 0.076 & 0.829 & 0.899 & 0.717 & \textbf{0.032} & \textbf{0.851} & 0.904 & 0.786 & 0.045 \\ %\cline{1-2}
 $f_3$+$f_5$   & 0.805 & 0.863 & 0.738 & 0.074 & 0.830 & 0.900 & 0.721 & 0.033 & 0.849 & 0.904 & 0.785 & 0.045 \\ %\cline{1-2}
 $f_2$+$f_5$ (Ours) & \textbf{0.812} & \textbf{0.870} & \textbf{0.749} & \textbf{0.073} & \textbf{0.831} & \textbf{0.901} & \textbf{0.722} & 0.033 & \textbf{0.851} & \textbf{0.907} & \textbf{0.788} & \textbf{0.044} \\ \hline
\end{tabular}}
\caption{Ablation studies of different inputs of EAM on three datasets. The best results are highlighted in \textbf{Bold}.}
\label{tab:EAM}
% \vspace{-1.5mm}
\end{table*}

\section{Experiments}
\label{sec:expe}

\subsection{Implementation Details}
We implement our model with PyTorch and employ Res2Net-50~\cite{res2net} pre-trained on ImageNet as our backbone. We resize all the input images to 416 x 416 and augment them by randomly horizontal flipping. During the training stage, the batch size is set to 16, and the Adam optimizer~\cite{adam} is adopted. The learning rate is initialized to 1e-4 and adjusted by poly strategy with the power of 0.9. Accelerated by an NVIDIA Tesla P40 GPU, the whole training takes about $\sim$2 hours with 25 epochs.

\subsection{Datasets} 
We evaluate our method on three public benchmark datasets: CAMO~\cite{camo}, COD10K~\cite{sinet} and NC4K~\cite{lsr}.
%\begin{itemize}
    %\item CAMO contains 1,250 camouflaged images (1,000 for training and 250 for testing) covering eight categories.
    %\item COD10K is currently the largest dataset, including 5066 camouflaged images (3,040 for training and 2,026 for testing) covering five super-classes and 69 sub-classes.
    %\item NC4K is currently the largest testing dataset with 4,121 images downloaded from the Internet.
%\end{itemize}
%\textbf{Training/Testing Sets}. We follow the previous works~\cite{sinet} to use the training set of CAMO and COD10K as our training set and others as our testing sets.
We follow the previous works~\cite{sinet}, which use the training set of CAMO and COD10K as our training set, and use their testing set and NC4K as our testing sets.

\subsection{Evaluation Metrics}
We utilize four widely used metrics to evaluate our method, \ie, mean absolute error (MAE, $\mathcal{M}$)~\cite{mae}, weighted F-measure ($F_\beta^w$)~\cite{fm}, structure-measure ($S_\alpha$)~\cite{sm} and mean E-measure ($E_\phi$)~\cite{em}.

\begin{figure}[bth]
    \centering{
	\includegraphics[width=0.9\linewidth]{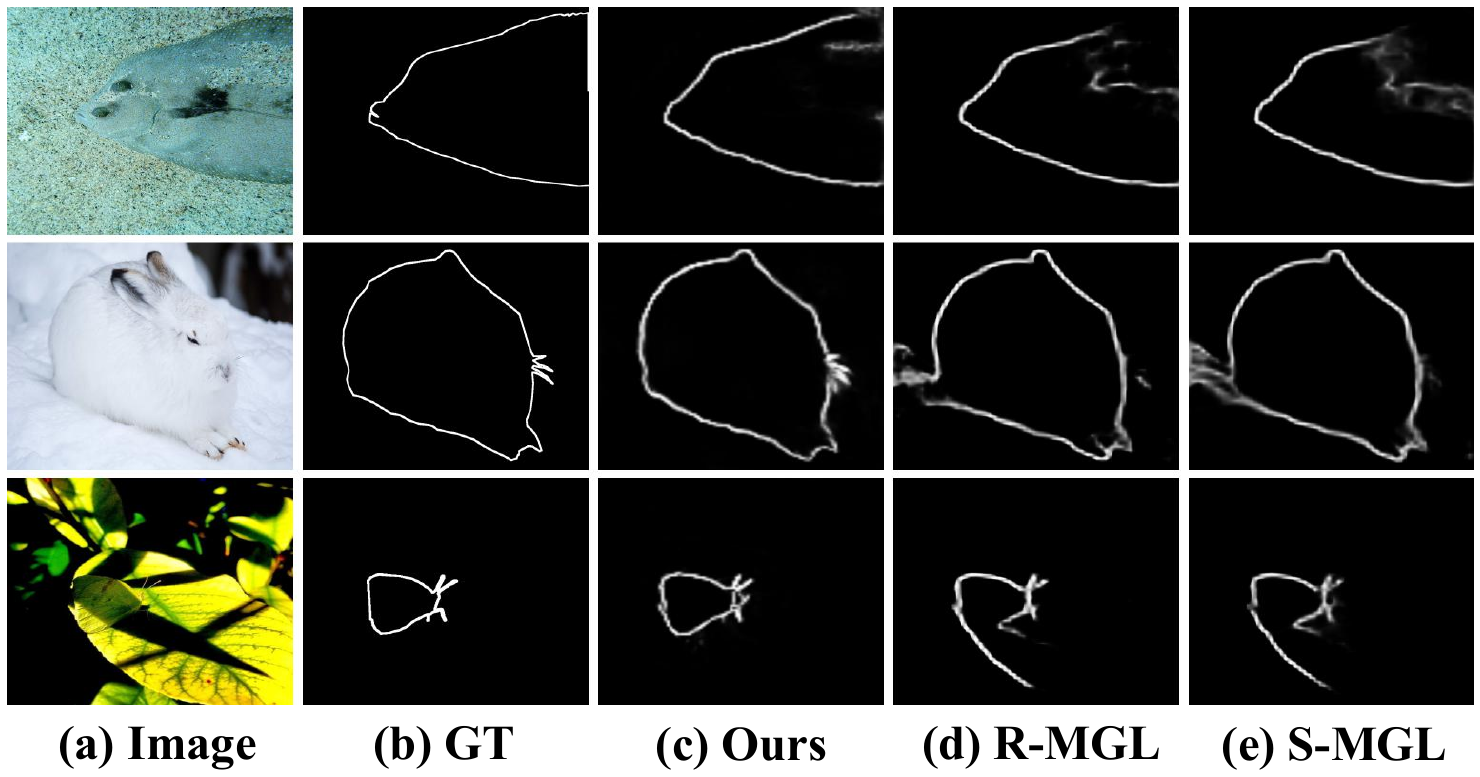}
	\caption{Examples of object-related edge exploration in EAM.} \label{fig:edge}
    }% \vspace{-4mm}
\end{figure}

\subsection{Comparison with State-of-the-arts}
To demonstrate the effectiveness of our method, we compare it against 18 state-of-the-art methods, including 10 SOD models, \ie, PoolNet~\cite{poolnet}, EGNet~\cite{zhao2019egnet}, SRCN~\cite{scrn}, F$^3$Net~\cite{f3net}, ITSD~\cite{ITSD}, CSNet~\cite{gao2020highly}, MINet~\cite{minet}, UCNet~\cite{zhang2020uc}, PraNet~\cite{pranet} and BASNet~\cite{bas}, and 8 COD models, \ie, SINet~\cite{sinet}, PFNet~\cite{pfnet}, S-MGL~\cite{mgl}, R-MGL~\cite{mgl}, LSR~\cite{lsr}, UGTR~\cite{ugtr}, C$^2$FNet~\cite{c2fnet} and JCSOD~\cite{jcsod}. For fair comparison, all the predictions of these methods are either provided by the authors or produced by models retrained with open source codes. % All results are evaluated with the same code.

\noindent\textbf{Quantitative Evaluation.}
Table~\ref{tab1} reports the quantitative results of our method against 18 competitors on three datasets. It is obvious that our method outperforms all other models on three datasets under four evaluation metrics. Specifically, compared with the second-best JCSOD, our method increases $S_\alpha$ by 1.80$\%$, $E_\phi$ by 1.40$\%$ and $F_\beta^w$ by 3.55$\%$ on average. Compared with the third-best C$^2$FNet, our method increases $S_\alpha$ by 1.93$\%$, $E_\phi$ by 1.41$\%$ and $F_\beta^w$ by 4.28$\%$ on average.

\noindent\textbf{Qualitative Evaluation}. 
Fig.~\ref{fig:results} shows the qualitative comparisons of different COD methods on several typical samples from the COD10K dataset, covering four super-classes, \ie,  aquatic, terrestrial, flying and, amphibious. These results intuitively show the superior performance of the proposed method. Noted that our method provides accurate camouflaged object predictions with finer and more complete object structure and boundary details.

\noindent\textbf{Boundary Exploration.} Fig.~\ref{fig:edge} shows the visual comparison of our model with MGL in terms of boundary-related edge extraction. It can be seen that, although MGL proposes an auxiliary edge detection network based on a complex graph model, it still loses many structural details, resulting in poor boundary localization in prediction. It is proven that our method achieves superior performance in object-related edge information mining and camouflaged object prediction.

\subsection{Ablation Study}
In order to validate the effectiveness of each key component, we design several ablation experiments and report the results in Tab.~\ref{tab:ablation}. 
%For baseline model (B), we remove all additional modules (\ie, EAM, EFM and CAM) and replace them with the standard convolutions for predictions.
For baseline model (B), we remove all the additional modules (\ie, EAM, EFM and CAM), and only retain the 1$\times$1 convolution in four EFMs to reduce the channels of the backbone features ($f_i, i=\{2,3,4,5\}$) and use the initial aggregation operation in the CAM to fuse the multi-level features in the top-down manner.

\noindent\textbf{Effectiveness of CAM}. From Tab.~\ref{tab:ablation}, compared with B model, the B+CAM model provides better performance. Especially, our module has more advantages on the metrics $F_{\beta}^{\omega}$ that shows 1.50\% performance increases averagely.

\noindent\textbf{Effectiveness of Edge Cues (EAM)}. To verify the effectiveness of object-related edge cues, we keep the initial fusion operation and final  1$\times$1 convolution in EFMs, and remove the local channel attention (LCA). From Tab.~\ref{tab:ablation}, the model c (B+EAM+EFM w/o LCA) achieves better overall performance compared with the baseline model a, especially in terms of $F_{\beta}^{\omega}$ with 1.15\% performance gains on average for all datasets. Thus, the edge prior extracted by EAM is beneficial to boost detection performance.

\noindent\textbf{Effectiveness of EFM}. 
Then we add LCA on model c, that is, the complete EFM, to validate effectiveness of the integration operation of edge cues and camouflaged object features. As can be seen in Tab.~\ref{tab:ablation}, the B+EAM+EFM model shows the performance improvement compared with model a and model c, demonstrating the effective contribution of LCA and the proposed EFM to the final predictions. 
Furthermore, combined with the designed EAM, EFM and CAM, the proposed BGNet achieves obvious performance improvements on all datasets, with the performance gains of 1.10\%, 1.14\% and 2.65\% on average in terms of $S_\alpha, E_\phi$ and $F_\beta^{\omega}$, respectively.

\noindent\textbf{Input of EAM}. We also test the effectiveness of different inputs for EAM, \eg, $f_1, f_2$ and $f_3$ used to explore edges with $f_5$ to help locate object-related edges. As shown in Tab.~\ref{tab:EAM}, the combination of $f_2+f_5$ obtains the best performance for camouflaged object detection.

\section{Conclusion}
\label{sec:con}

In this paper, we resort to edge priors to assist recovering object structure and boost the performance of camouflaged object detection. We propose a simple yet effective boundary-guided network (\textit{BGNet}), which contains edge-aware module, edge-guidance feature module, and context aggregation module, to explore object-related edge semantics to guide and enhance representation learning for COD. By adopting edge cues, our \textit{BGNet} provides accurate camouflaged object predictions with complete and fine object structure and boundaries. Extensive experiments show that our method outperforms existing state-of-the-art methods on three benchmarks.

% \newpage

%% The file named.bst is a bibliography style file for BibTeX 0.99c
\bibliographystyle{named}
\bibliography{BGNet-arxiv}

\newpage

\section{Appendix}

\subsection{Model complexity}

Our \textit{BGNet} is a multi-task learning model for COD, so we report the comparison of existing COD methods based on multi-task learning (\textit{i.e.}, S-MGL, R-MGL, JCSOD, LSR) with ours in terms of number of parameters, FLOPs and FPS in Tab.~\ref{tab:complexity}. All evaluations follow the inference settings in the corresponding papers. It can be seen that our method achieves a good balance of these three metrics.

\begin{table}[h]
\centering
\begin{tabular}{l|cccc}
\toprule
Method & Input Size  & Params.~$\downarrow$  & FLOPs~$\downarrow$ & FPS~$\uparrow$ \\ 
\midrule
S-MGL  &473$\times$473  &63.60M  &236.60G  &12\\
R-MGL  &473$\times$473  &67.64M  &249.89G  &8\\
JCSOD  &352$\times$352  &121.63M  &25.20G  &43\\
LSR    &352$\times$352  &57.90M  &25.21G  &33\\
BGNet  &416$\times$416  &79.85M  &58.45G  &35\\
\bottomrule
\end{tabular}
\caption{Comparison of model complexity.}
\label{tab:complexity}
\end{table}

\subsection{Experimental Settings for Ablation Study}

Figure~\ref{fig:net_eam} shows the specific experimental settings for model (b) (B+EAM+EFM w/o LCA) and model (c) (B+EAM+EFM) in Tab.~\ref{tab:ablation}. Comparing the results of model (b) and model (c), that is, B+EAM+EFM (w/o LCA) and B+EAM+EFM(w/ LCA), it can be found that the boundary cues are still effective even if they are fused in a simple way, while local attention is the icing on the cake.

\begin{figure}[h]
    \centering
	\includegraphics[width=1\linewidth]{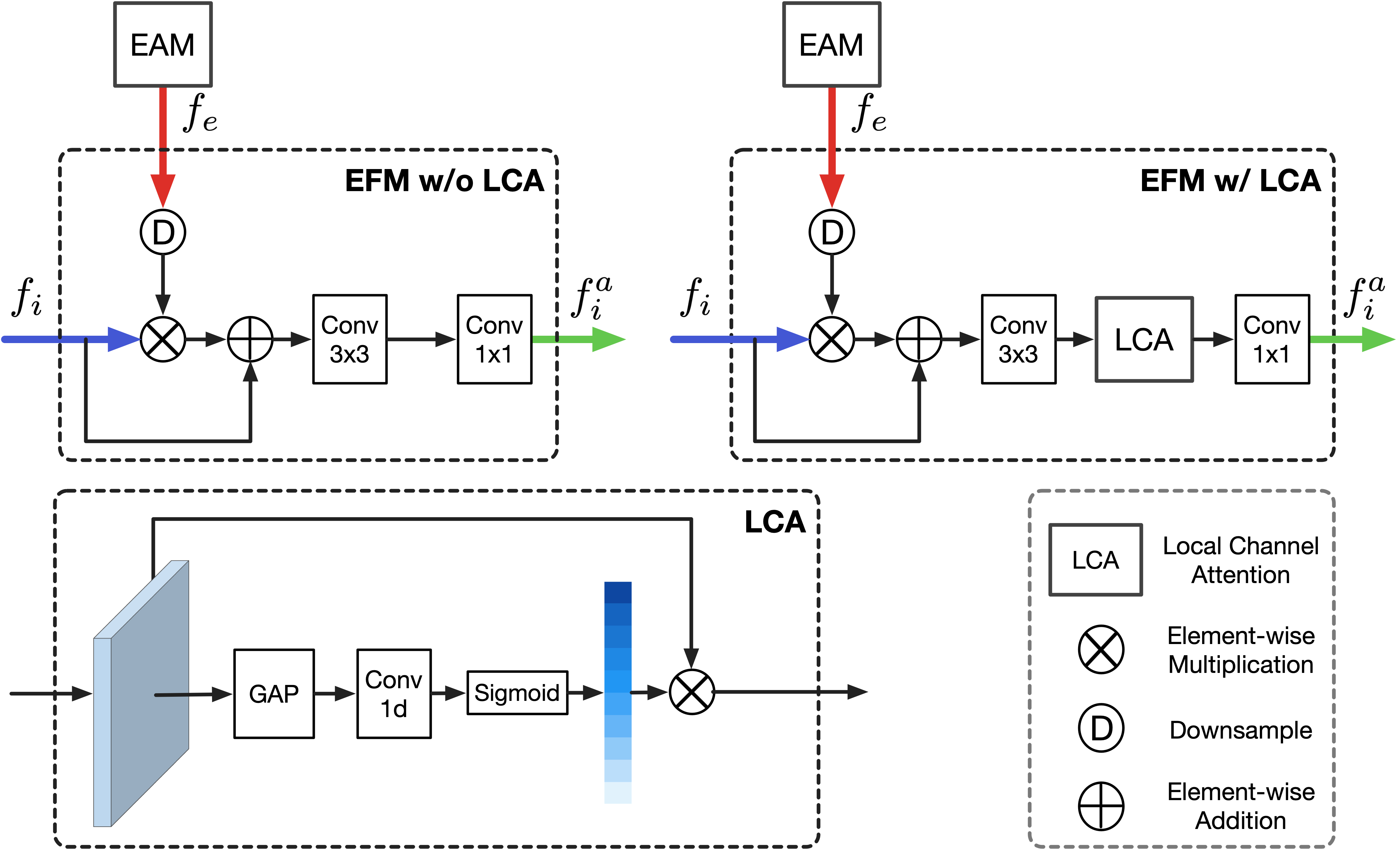}
	\caption{Illustration of the edge-guidance feature module (EFM) with and without local channel attention (LCA) mechanism.}
    \label{fig:net_eam}
\end{figure}

\subsection{Parameter $\lambda$} 

Our \textit{BGNet} has three mask predictions and one boundary prediction. Intuitively, we set $\lambda$ to 3 to balance these two types of losses in the total loss. We also study the different parameter settings (\textit{i.e.}, $\lambda=\{1,2,3,4,5\}$) in Tab.~\ref{tab:lamda}. Four metrics (\textit{i.e.}, $E_\phi$, $F_\beta^w$, $S_\alpha$, $\mathcal{M}$) are adopted for performance evaluation. From these results, we can see that $\lambda=3$ achieves the best performance compared to other settings.

\begin{table}[h]
\centering
\begin{tabular}{l|llll}
\toprule
$\lambda$  & $E_\phi\uparrow$ & $F_\beta^w\uparrow$   & $S_\alpha\uparrow$   & $\mathcal{M}\downarrow$ \\ 
\midrule
$\lambda=1$  &0.862    &0.743  &0.809  &0.073\\
$\lambda=2$  &0.865    &0.742  &0.806  &0.073\\
$\lambda=3$  &\textbf{0.869}  &\textbf{0.748}  &\textbf{0.811}  &0.073 \\
$\lambda=4$  &0.863    &0.736  &0.803  &\textbf{0.072}\\
$\lambda=5$  &0.867    &0.739  &0.806  &0.074\\
\bottomrule
\end{tabular}
\caption{Quantitative results for different setting of $\lambda$ on CAMO.}
\label{tab:lamda}
\end{table}

% Our original intention of setting $\lambda$ to 3 is to balance the coefficients of the two types of losses in the total loss. In addition, we conducted five sets of experiments on the value of $\lambda$ ($\lambda={1,2,3,4,5}$). According to the experimental results, we find that two larger datasets, COD10K-Test and NC4K, are not sensitive to the value of $\lambda$, so we only report the performance of different $\lambda$ values on the smaller CAMO-Test in Table~\ref{tab:lamda}. As shown in Table~\ref{tab:lamda}, $\lambda=3$ outperforms other settings on all five evaluation metrics.

\subsection{Relationship between modules} 

Our \textit{BGNet} is a unified whole with serially interdependent components: EAM, EFM, and CAM, which complement each other. The former two explore boundary cues and guide the feature learning, and then the latter further enhances critical feature representation. As shown in Tab.~\ref{tab:ablation} of the paper, the contributions of EAM, EFM and CAM are 1.22\%, 0.65\%, 1.94\% on average respectively in terms of $F_\beta^{w}$. Thus, we can see that CAM and EAM (boundary cues) contribute more than EFM.

\end{document}